\pdfoutput=1

\documentclass[11pt]{article}

\usepackage{acl}

\usepackage{times}
\usepackage{latexsym}
\usepackage{colortbl}
\usepackage[T1]{fontenc}

\usepackage[utf8]{inputenc}

\usepackage{microtype}

\usepackage{inconsolata}

\usepackage[ruled]{algorithm2e}
\usepackage{algorithmic} 
\usepackage{listings}
\usepackage{longtable}
\usepackage{soul}
\usepackage[utf8]{inputenc} 
\usepackage[T1]{fontenc}    
\usepackage{hyperref}       
\usepackage{url}            
\usepackage{booktabs}       
\usepackage{amsfonts}       
\usepackage{nicefrac}       
\usepackage{microtype}      
\usepackage{xcolor}         
\usepackage{natbib}
\usepackage{comment}
\usepackage{fancyvrb}
\usepackage{enumitem} 
\usepackage{python}
\usepackage{graphicx}
\usepackage{amsmath}
\usepackage{amssymb}
\usepackage{mathtools}
\usepackage{amsthm}
\usepackage{subcaption}
\usepackage{tabularx}
\usepackage{pgfplots}
\usetikzlibrary{pgfplots.groupplots}
\usepackage{wrapfig, lipsum, booktabs}
\pgfplotsset{compat=1.3}
\usepackage{tikz}
\usepackage{bm}
\usepackage{caption}
\usepackage{multirow}
\usepackage{comment}
\usepackage{color}
\usepackage[group-separator={,},group-minimum-digits={3}]{siunitx}
\graphicspath{ {./} }
\UseRawInputEncoding
\usepackage{xspace} 
\usepackage{textcomp}
\usepackage{hhline}
\usepackage{bbding}
\usepackage{makecell}
\usepackage{arydshln}
\usepackage{vcell}
\usepackage{pifont}
\usepackage{fontawesome}
\usepackage{wasysym}
\newcommand{\okmark}{{\textbf{\textcolor[rgb]{0.1, 0.5, 0.1}{$\checkmark$}}}}
\newcommand{\ngmark}{{\textbf{\color{red}{\ding{55}}}}}
\usepackage{inconsolata}
\definecolor{myblue}{RGB}{215,238,247}
\definecolor{mygreen}{RGB}{230,241,221}
\definecolor{mygrey}{RGB}{242,242,242}

\newcommand{\cappyicon}{\includegraphics[scale=0.016]{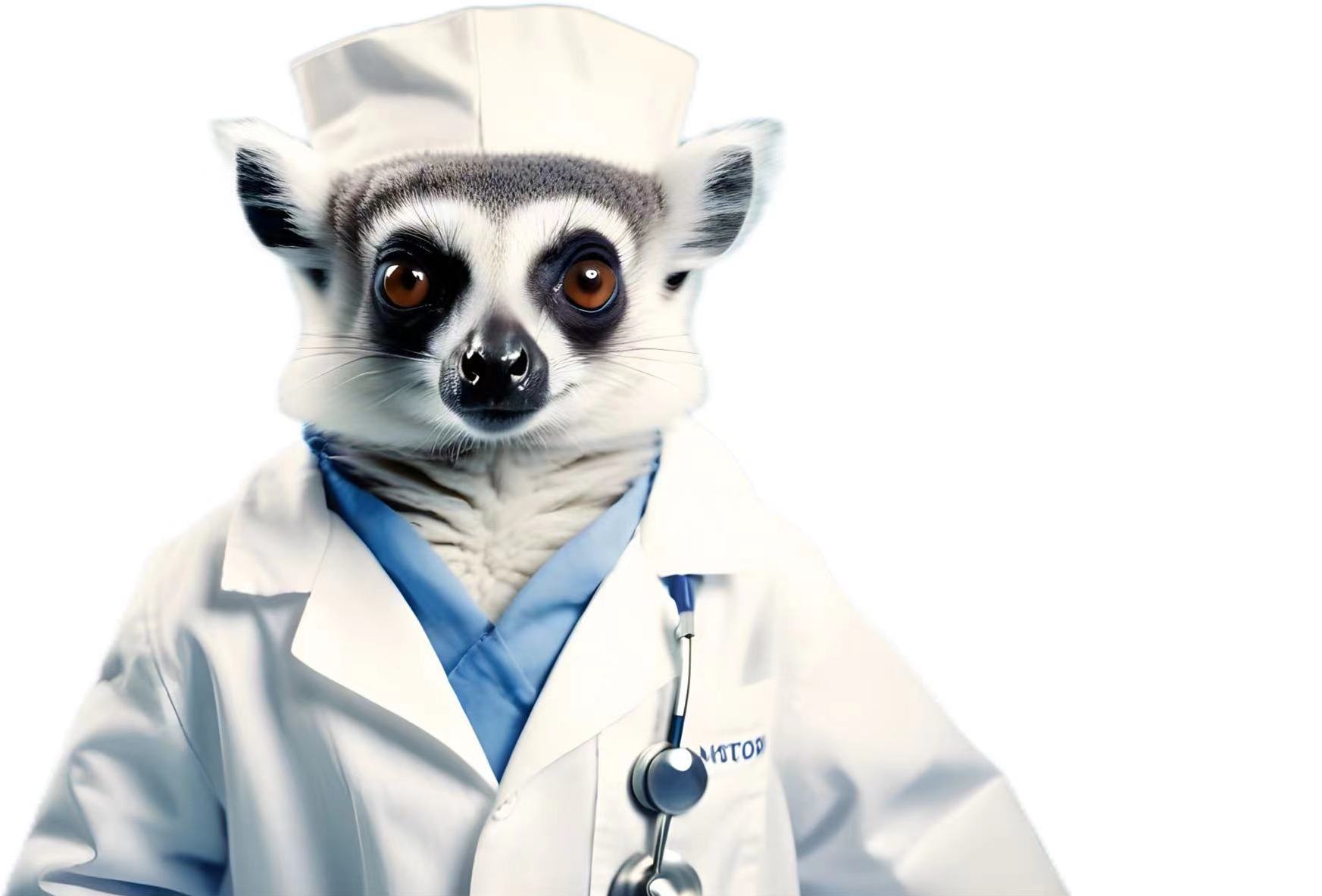}}

\title{\cappyicon  \textsc{MedAgents:} Large Language Models as Collaborators for Zero-shot Medical Reasoning}

\author{
  Xiangru Tang$^{\heartsuit}$\thanks{\quad Equal contribution.} , 
  Anni Zou$^{\spadesuit}$\footnotemark[1] , 
  Zhuosheng Zhang$^{\spadesuit}$,
  Ziming Li$^\heartsuit$,\\
  \textbf{
  Yilun Zhao$^{\heartsuit}$,
  Xingyao Zhang$^\heartsuit$,
  Arman Cohan$^\heartsuit$,
  Mark Gerstein$^\heartsuit$}\\
  $^\heartsuit$Yale University  \quad
  $^\spadesuit$Shanghai Jiao Tong University \\
 \texttt{ 
xiangru.tang@yale.edu, mark@gersteinlab.org
  }\\
}

\begin{document}
\maketitle
\begin{abstract}
Large language models (LLMs), despite their remarkable progress across various general domains, encounter significant barriers in medicine and healthcare. This field faces unique challenges such as domain-specific terminologies and reasoning over specialized knowledge. To address these issues, we propose \textsc{MedAgents}, a novel multi-disciplinary collaboration framework for the medical domain. MedAgents leverages LLM-based agents in a role-playing setting that participate in a collaborative multi-round discussion, thereby enhancing LLM proficiency and reasoning capabilities. This training-free framework encompasses five critical steps: gathering domain experts, proposing individual analyses, summarising these analyses into a report, iterating over discussions until a consensus is reached, and ultimately making a decision. Our work focuses on the zero-shot setting, which is applicable in real-world scenarios. Experimental results on nine datasets (MedQA, MedMCQA, PubMedQA, and six subtasks from MMLU) establish that our proposed \textsc{MedAgents} framework excels at mining and harnessing the medical expertise within LLMs, as well as extending its reasoning abilities. 
Our code can be found at \url{https://github.com/gersteinlab/MedAgents}.
\end{abstract}

\section{Introduction}
Large language models (LLMs) \citep{few-shot, scao2022bloom, chowdhery2022palm, touvron2023llama, openai2023gpt4} have exhibited notable generalization abilities across a wide range of tasks and applications \citep{lu2023chameleon, zhou2023webarena, Park2023GenerativeAgents}, with these capabilities stemming from their extensive training on vast comprehensive corpora covering diverse topics. 
However, in real-world scenarios, LLMs tend to encounter domain-specific tasks that necessitate a combination of domain expertise and complex reasoning abilities \citep{moor2023foundation, wu2023precedentenhanced, clinical, yang2023investlm}. Amidst this backdrop, a noteworthy research topic lies in the adoption of LLMs in the medical field, which has gained increasing prominence recently \citep{zhang2023alpacareinstructiontuned, bao2023discmedllm, clinical}.

Two major challenges prevent LLMs from effectively handling tasks in the medical sphere:
(i) Limited \emph{volume and specificity} of medical training data compared to the vast general text data, due to cost and privacy concerns \citep{thirunavukarasu2023large}.\footnote{Although Med-PaLM 2 \citep{singhal2023towards} serves as a specialized medical LLM fine-tuned on the basis of PaLM 2, it is closed-sourced and not publicly accessible yet.}
(ii) The demand for \textit{extensive domain knowledge} \citep{schmidt2007expertise} and \textit{advanced reasoning skills} \citep{lievin2022can} makes eliciting medical expertise via simple prompting challenging \citep{kung2023performance, clinical}.
Although numerous attempts, particularly within math and coding, have been made to enhance prompting methods \citep{besta2023graph, lala2023paperqa,wang2023augmenting}, strategies used in the medical field have been shown to induce \emph{hallucination} \citep{umapathi2023med, harris2023large,ji2023survey}, indicating the need for more robust approaches.

\begin{figure*}[htb]
    \centering
    \includegraphics[width=0.98\textwidth]{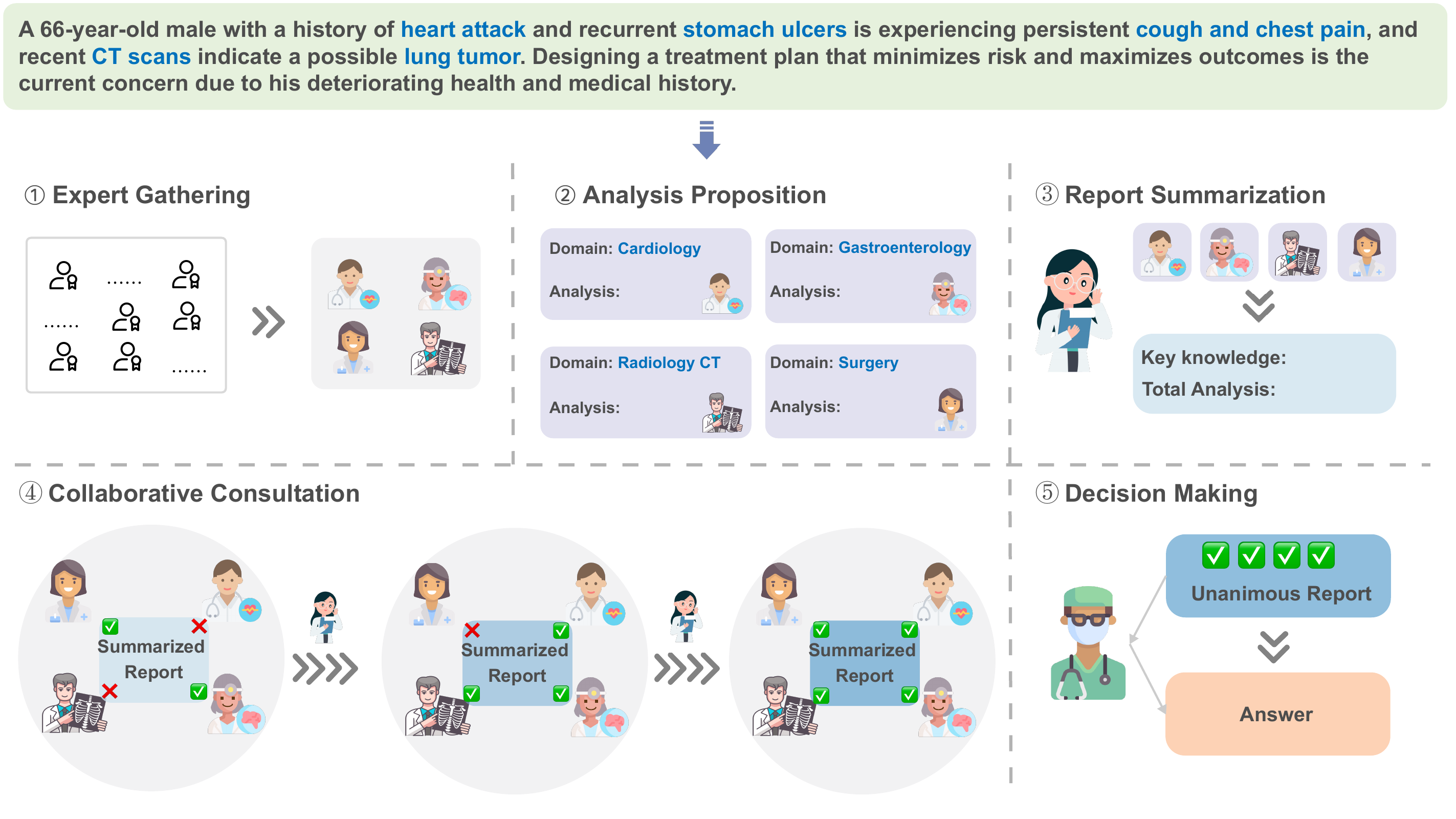}
    \caption{Diagram of our proposed \textsc{MedAgents} framework. Given a medical question as input, the framework performs reasoning in five stages: (i) expert gathering, (ii) analysis proposition, (iii) report summarization, (iv) collaborative consultation, and (v) decision making.}
    \label{fig:overview}
    \vspace{-.4cm}
\end{figure*}

Meanwhile, recent research has surprisingly witnessed the success of multi-agent collaboration \citep{xi2023rise, wang2023unleashing} which highlights the simulation of human activities \citep{du2023improving, liang2023encouraging, Park2023GenerativeAgents} and optimizes the collective power of multiple agents \citep{chen2023agentverse, li2023metaagents, hong2023metagpt}. Through such design, the expertise implicitly embedded within LLMs, or that the model has encountered during its training, which may not be readily accessible via traditional prompting, is effectively brought to the fore. This process subsequently enhances the model's reasoning capabilities throughout multiple rounds of interaction \citep{wang2023rolellm, fu2023improving}.


Motivated by these notions, we pioneer a \textbf{multi-disciplinary collaboration framework (MedAgents)} specifically tailored to the clinical domain. 
Our objective centers on unveiling the intrinsic medical knowledge embedded in LLMs and reinforcing reasoning proficiency in a training-free manner. 
As shown in Figure \ref{fig:overview}, the \textsc{MedAgents} framework gathers experts from diverse disciplines and reaches consistent conclusions through collaborative discussions.

Based on our \textsc{MedAgents} framework, we conduct experiments on nine datasets, including MedQA~\citep{jin2021disease}, MedMCQA~\citep{pal2022medmcqa}, PubMedQA~\citep{jin2019pubmedqa} and six medical subtasks from MMLU~\citep{hendrycks2020measuring}.\footnote{We follow the evaluation setting from Faln-PaLM \citep{clinical}.} To better align with real-world application scenarios, our study focuses on the zero-shot setting, which can serve as a plug-and-play method to supplement existing medical LLMs such as Med-PaLM 2 \citep{singhal2023towards}.
Encouragingly, our proposed approach outperforms settings for both chain-of-thought (CoT) and self-consistency (SC) prompting methods.
Most notably, our approach achieves better performance under the zero-shot setting compared with the 5-shot baselines.


Based on our results, we further investigate the influence of agent numbers and conduct human evaluations to pinpoint the limitations and issues prevalent in our approach. We find four common categories of errors: (i) lack of domain knowledge, (ii) mis-retrieval of domain knowledge, (iii) consistency errors, and (iv) CoT errors.
Targeted refinements focused on mitigating these particular shortcomings would enhance the model's proficiency and reliability.

\begin{figure*}[h]
    \centering
    \includegraphics[width=1.0\textwidth]{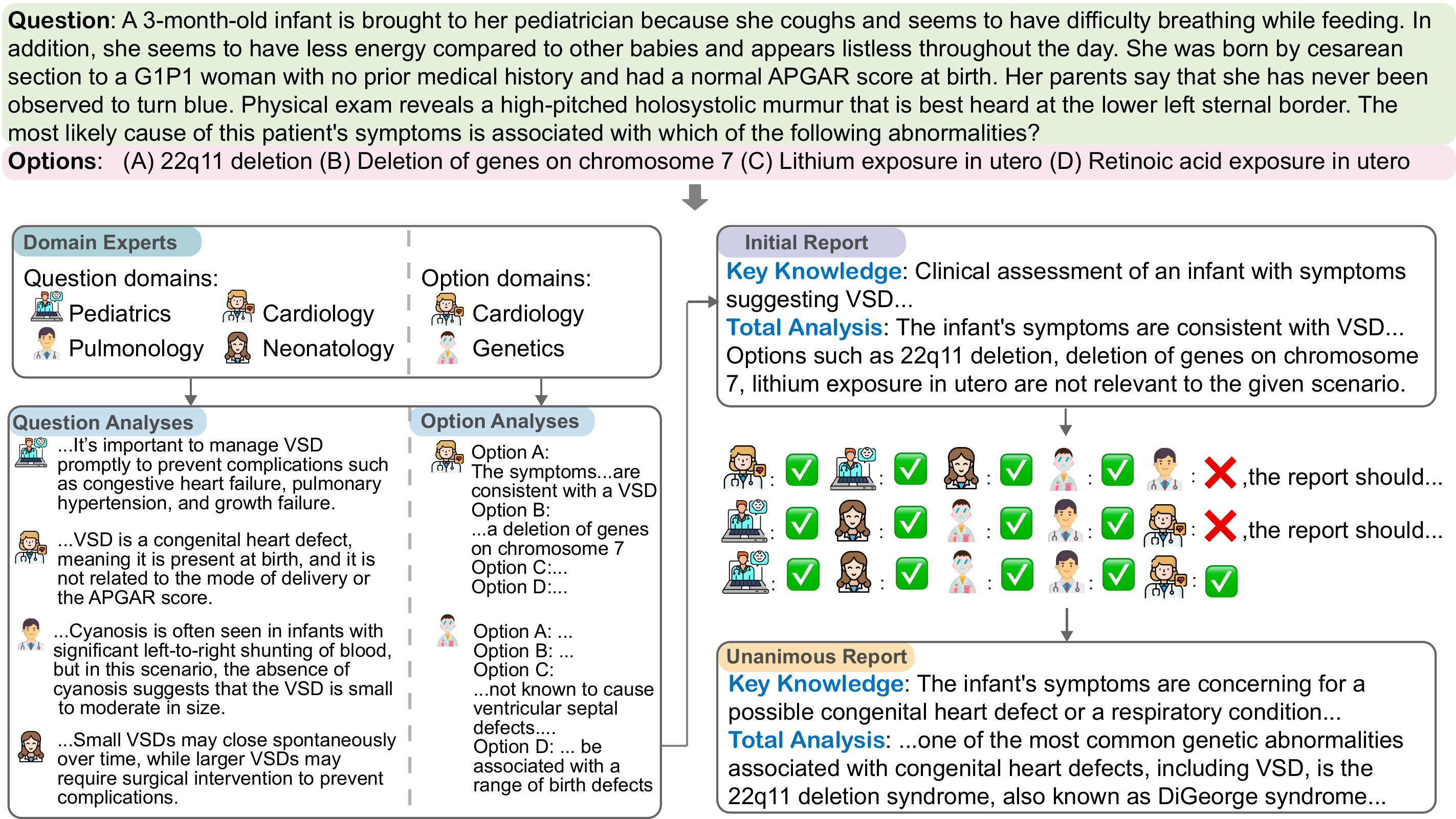}
    \caption{Illustrative example of our proposed MedAgents, a multi-disciplinary collaboration framework. The questions and options are first presented, with domain experts subsequently gathered. The recruited experts conduct thorough Question and Option analyses based on their respective fields. An initial report synthesizing these analyses is then prepared to concisely represent the performed evaluations. The assembled LLM experts, possessing respective disciplinary backgrounds, engage in a discussion over the initial report, voicing agreements and disagreements. Ultimately, after iterative refinement and consultation, a unanimous report is generated that best represents the collective expert knowledge and reasoning on the given medical problem.
    }
    \label{fig:example}
    \vspace{-.4cm}
\end{figure*}

Our contributions are summarized as follows: 

(i) To the best of our knowledge, we are the first to propose a multi-agent framework within the medical domain and explore how multi-agent communication within the medical setting can lead to a consensus decision, adding a novel dimension to the current literature on medical question answering.

(ii) Our proposed \textsc{MedAgents} framework enjoys enhanced faithfulness and interpretability by harnessing role-playing and collaborative agent discussion. And we demonstrate that role-playing allows LLM to explicitly reason with accurate knowledge, without the need for retrieval augmented generation (RAG).
Examples illustrating interpretability are shown in Appendix \ref{app:case}.

(iii) Experimental results on nine datasets demonstrate the general effectiveness of our proposed \textsc{MedAgents} framework. Besides, we identify and categorize common error types in our approach through rigorous human evaluation to shed light on future studies.


\section{Method}

This section presents the details of our proposed multi-disciplinary collaboration \textsc{MedAgents} framework. Figure \ref{fig:overview} and \ref{fig:example} give an overview and an illustrative example of its pipeline, respectively. 
Our proposed \textsc{MedAgents} framework works in five stages: 
(i) expert gathering: assemble experts from various disciplines based on the clinical question;
(ii) analysis proposition: domain experts present their own analyses with their expertise;
(iii) report summarization: develop a report summary on the basis of previous analyses;
(iv) collaborative consultation: hold a consultation over the summarized report with the experts. The report will be revised repeatedly until every expert has given their approval.
(v) decision making: derive a final decision from the unanimous report.\footnote{Details about all guideline prompts and roles are shown in Section \ref{app:prompt} for clarification.}

\subsection{Expert Gathering}
Given a clinical question $q$ and a set of options $op = \left\{o_1, o_2, \ldots, o_k\right\}$ where $k$ is the number of options, the goal of the Expert Gathering stage is to recruit a group of question domain experts $\mathcal{QD} = \left\{qd_1, qd_2, \ldots, qd_m\right\}$ and option domain experts $\mathcal{OD} = \left\{od_1, od_2, \ldots, od_n\right\}$ where $m$ and $n$ represent the number of question domain experts and option domain experts, respectively. \footnote{We design domain experts for the question and options in order to leverage diverse agents to elicit multifaceted and comprehensive knowledge.}
Specifically, we assign a role to the model and provide instructions to guide the model output to the corresponding domains based on the input question and options, respectively: 
\begin{equation}
    \begin{split}
        \mathcal{QD} &= \text{LLM}\left(q, \texttt{r}_\texttt{qd}, \texttt{prompt}_\texttt{qd}\right), \\
        \mathcal{OD} &= \text{LLM}\left(q, op, \texttt{r}_\texttt{od}, \texttt{prompt}_\texttt{od}\right), \\
    \end{split}
    \label{expert}
\end{equation}
where $\left(\texttt{r}_\texttt{qd}, \texttt{prompt}_\texttt{qd}\right)$ and $\left(\texttt{r}_\texttt{od}, \texttt{prompt}_\texttt{od}\right)$ stand for the system role and guideline prompt to gather domain experts for the question $q$ and options $op$.


\begin{table*}[t]
    \centering
    \setlength{\tabcolsep}{4pt}
    \small
    \begin{tabular}{lcccc}
        \toprule
    \textbf{Dataset} & \textbf{Format} & \textbf{Choice} & \textbf{Testing Size} & \textbf{Domain} \\ 
        \midrule
    MedQA & Question + Answer & A/B/C/D & 1273 &  US Medical Licensing Examination \\ 
        \midrule
    MedMCQA & Question + Answer & A/B/C/D and Explanations & 6.1K &  AIIMS and NEET PG entrance exams \\ 
        \midrule
    PubMedQA & Question + Context + Answer & Yes/No/Maybe & 500 & PubMed paper abstracts \\ 
        \midrule
    \multirow{2}{*}{MMLU} &  \multirow{2}{*}{Question + Answer} & \multirow{2}{*}{ A/B/C/D}  & \multirow{2}{*}{1089} &Graduate Record Examination  \\     & &  & &   \& US Medical Licensing Examination \\ 
        \bottomrule
\end{tabular}
    \caption{Summary of the Datasets. Part of the values are from the appendix of \cite{clinical}.}\label{table:datasets}
\vspace{-.4cm}
\end{table*}

\subsection{Analysis Proposition}
After gathering domain experts for the question $q$ and options $op$, we aim to inquire experts to generate corresponding analyses prepared for later reasoning:  $\mathcal{QA} = \{qa_1, qa_2, \ldots, qa_m\}$ and $\mathcal{OA} = \{oa_1, oa_2, \ldots, oa_n\}$.

\paragraph{Question Analyses} \label{sec:qa}
Given a question $q$ and a question domain $qd_i \in \mathcal{QD}$, we ask LLM to serve as an expert specialized in domain $qd_i$ and derive the analyses for the question $q$ following the guideline prompt $\texttt{prompt}_\texttt{qa}$:
\begin{equation}
    qa_i = \text{LLM}\left(q, qd_i, \texttt{r}_\texttt{qa}, \texttt{prompt}_\texttt{qa}\right).
    \label{q_anal}
\end{equation}


\paragraph{Option Analyses}
Now that we have an option domain $od_i$ and question analyses $\mathcal{QA}$, we can further analyze the options by taking into account both the relationship between the options and the relationship between the options and question. Concretely, we deliver the question $q$, the options $op$, a specific option domain $od_i \in \mathcal{OD}$, and the question analyses $\mathcal{QA}$ to the LLM: 
\begin{equation}
    oa_i = \text{LLM}\left(q, op, od_i, \mathcal{QA}, \texttt{r}_\texttt{oa}, \texttt{prompt}_\texttt{oa}\right).
    \label{o_anal}
\end{equation}

\subsection{Report Summarization}
In the Report Summarization stage, we attempt to summarize and synthesize previous analyses from various domain experts $\mathcal{QA} \cup \mathcal{OA}$. 
Given question analyses $\mathcal{QA}$ and option analyses $\mathcal{OA}$, we ask LLMs to play the role of a medical report assistant, allowing it to generate a synthesized report by extracting key knowledge and total analysis based on previous analyses: 
\begin{equation}
    Repo = \text{LLM}\left(\mathcal{QA}, \mathcal{OA}, \texttt{r}_\texttt{rs}, \texttt{prompt}_\texttt{rs}\right).
    \label{rep_summ}
\end{equation}

\vspace{-.1cm}

\begin{algorithm}[h]
\small
\caption{Collaborative Consultation}\label{Algo}
\KwIn{Domain experts $D = \{d_1,...,d_n\}$, initial report $R_0$, Model $\mathcal{M}$, maximum attempts $t$, prompts $\{p_{vote}, p_{mod}, p_{rev}\}$ }
\KwOut{Final report $R_{f}$}
\BlankLine
\tcp{Initialize variables}
$nocon\_flag\leftarrow True$, $n_{try}\leftarrow 0$ \\
$R_{cur}\leftarrow R_0$, $Mods\leftarrow \emptyset$ \\

\BlankLine
\tcp{Iterative review}
\While{$nocon\_flag$ is $True$ and $n_{try} < t$}{
        $n_{try} \leftarrow n_{try} +1$ \\
        $nocon\_flag\leftarrow False$ \\
        \tcp{vote for the report}
        \For{$i$ in $1,...,n$}{
            $vote_i \leftarrow \mathcal{M}(R_{cur}, d_i, p_{vote})$ \\
            \tcp{propose modifications}
            \If{$vote_i$ is $no$}{
                $Mod_i \leftarrow \mathcal{M}(R_{cur}, d_i, p_{mod})$ \\
                Update $Mods$ with $Mod_i$ \\
                $nocon\_flag\leftarrow True$ \\
            }
        }
        \tcp{modify the report}
        \If{$nocon\_flag$ is $True$}{
            $R_{cur} \leftarrow \mathcal{M}(R_{cur}, Mods, p_{rev})$ \\
        }
}
\Return{$R_f\leftarrow R_{cur}$}
\end{algorithm}


\subsection{Collaborative Consultation}

Since we have a preliminary summary report $Repo$, the objective of the Collaborative Consultation stage is to engage distinct domain experts in multiple rounds of discussions and ultimately render a summary report that is recognized by all experts. The overall procedure of this phase is presented in Algorithm \ref{Algo}. During each round of discussions, the experts give their votes (\emph{yes}/\emph{no}): $vote = \text{LLM}\left(Repo, \texttt{r}_\texttt{vote}, \texttt{prompt}_\texttt{vote}\right)$, as well as modification opinions if they vote \emph{no} for the current report. Afterward, the report will be revised based on the modification opinions. Specifically, during the $j$-th round of discussion, we note the modification comments from the experts as $Mod_j$, then we can acquire the updated report as $Repo_j = \text{LLM}\left(Repo_{j-1}, Mod_j, \texttt{prompt}_\texttt{mod}\right)$.
In this way, the discussions are held iteratively until all experts vote \emph{yes} for the final report $Repo_f$ or the discussion number attains the maximum attempts threshold.

\subsection{Decision Making}
In the end, we demand LLM act as a medical decision maker to derive the final answer to the clinical question $q$ referring to the unanimous report $Repo_f$:
\begin{equation}
    ans = \text{LLM}\left(q, op, Repo_f, \texttt{prompt}_\texttt{dm}\right).
    \label{deci}
\end{equation}

\vspace{-.3cm}

\begin{table*}[t]
\centering
\setlength{\tabcolsep}{2pt}
\small
\begin{tabular}{lcccccccccc}
\toprule
\multirow{2}{*}{\textbf{Method}} &  \multirow{2}{*}{MedQA} & \multirow{2}{*}{MedMCQA} & \multirow{2}{*}{PubMedQA}  & \multirow{2}{*}{Anatomy}&Clinical & College & Medical & Professional & College & \multirow{2}{*}{Avg.}\\

&  &  &  & & knowledge & medicine & genetics & medicine & biology & \\ 
\midrule  \\[-1em]
\rowcolor{gray!25} \multicolumn{11}{c}{\textbf{\texttt{~~~~~~Flan-Palm}}}  \\
Few-shot CoT    & 60.3&53.6& 77.2&66.7&77.0&83.3&75.0&76.5&71.1 &71.2\\
Few-shot CoT + SC   & 67.6&57.6&75.2&71.9&80.4&88.9&74.0&83.5&76.3 &75.0\\

\midrule\midrule  \\[-1em]

\rowcolor{gray!25} \multicolumn{11}{c}{\textbf{\texttt{~~~~~~GPT-3.5}}}  \\
\multicolumn{10}{l}{\textit{*few-shot setting}}  \\
Few-shot    &54.7 &56.7 &67.6 &65.9 &71.3 &59.0 &72.0 &75.7 &73.6 &66.3 \\
Few-shot CoT    &55.3 &54.7 &71.4 &48.1 &65.7 &55.5 &57.0 &69.5 &61.1 &59.8\\
Few-shot CoT + SC   &62.1 &58.3 &73.4 &70.4 &76.2 &69.8 &78.0 &79.0 &77.2 &71.6\\
\midrule
\multicolumn{10}{l}{\textit{*zero-shot setting}}  \\
Zero-shot       &54.3 &56.3 &73.7 &61.5 &76.2 &63.6 &74.0 &75.4 &75.0 &67.8\\
Zero-shot CoT   &44.3 &47.3 &61.3 &63.7 &61.9 &53.2 &66.0 &62.1 &65.3 &58.3\\
Zero-shot CoT + SC  &61.3 &52.5 &\textbf{75.7} &\textbf{71.1} &75.1 &68.8 &76.0 &\textbf{82.3} &75.7 &70.9\\
\sethlcolor{myblue}\hl{\texttt{\textsc{MedAgents}} (\textbf{Ours})}&\textbf{64.1}&\textbf{59.3}&72.9&65.2&\textbf{77.7}&\textbf{69.8}&\textbf{79.0}&82.1&\textbf{78.5} &\textbf{72.1}\\

\midrule\midrule  \\[-1em]

\rowcolor{gray!25} \multicolumn{11}{c}{\textbf{\texttt{~~~~~~GPT-4}}}  \\
\multicolumn{10}{l}{\textit{*few-shot setting}}  \\
Few-shot            &76.6 &70.1 &73.4 &79.3 &89.5 &75.6 &\textbf{93.0} &91.5 &91.7 &82.3 \\
Few-shot CoT        &73.3 &63.2 &74.9 &75.6 &89.9 &61.0 &79.0 &79.8 &63.2 &73.3\\
Few-shot CoT + SC   &82.9 &73.1 &75.6 &80.7 &90.0 &\textbf{88.2} &90.0 &95.2 &93.0 &85.4\\
\midrule
\multicolumn{10}{l}{\textit{*zero-shot setting}}  \\
Zero-shot           &73.0 &69.0 &76.2 &78.5 &83.3 &75.6 &90.0 &90.0 &90.0 &80.6\\
Zero-shot CoT       &61.8 &69.0 &71.0 &82.1 &85.2 &80.8 &92.0 &93.5 &91.7 &80.8\\
Zero-shot CoT + SC  &74.5 &70.1 &75.3 &80.0 &86.3 &81.2 &\textbf{93.0} &94.8 &91.7 &83.0 \\
\sethlcolor{myblue}\hl{\texttt{\textsc{MedAgents}} (\textbf{Ours})}&\textbf{83.7}&\textbf{74.8}&\textbf{76.8}&\textbf{83.5}&\textbf{91.0}&87.6&\textbf{93.0}&\textbf{96.0}&\textbf{94.3}&\textbf{86.7}\\

\bottomrule
\end{tabular}
\vspace{-.1cm}
\caption{Main results (Acc). SC denotes the self-consistency prompting method. Results in \textbf{bold} are the best performances.}\label{tab:exp_result1}
\vspace{-.2cm}
\end{table*}

\section{Experiments}
\subsection{Setup}\label{sec:ex_setup}
\paragraph{Tasks and Datasets.}
We evaluate our \textsc{MedAgents} framework on three benchmark datasets MedQA~\citep{jin2021disease}, MedMCQA~\citep{pal2022medmcqa}, and PubMedQA~\citep{jin2019pubmedqa}, as well as six subtasks most relevant to the medical domain from MMLU datasets ~\citep{hendrycks2020measuring} including anatomy, clinical knowledge, college medicine, medical genetics, professional medicine, and college biology. Table \ref{table:datasets} summarizes the data statistics. More information about the evaluated datasets is presented in Appendix \ref{app:dataset}.
\paragraph{Implementation.}
We utilize the popular and publicly available GPT-3.5-Turbo and GPT-4 \citep{openai2023gpt4} from Azure OpenAI Service.\footnote{\url{https://learn.microsoft.com/en-us/azure/ai-services/openai/}}
All experiments are conducted in the \textbf{zero-shot} setting. The temperature is set to 1.0 and \emph{top\_p} to 1.0 for all generations. The iteration number and temperature of SC are 5 and 0.7, respectively. The number $k$ of options is 4 except for PubMedQA (3). The numbers of domain experts for the question and options are set as: $m=5, n=2$ except for PubMedQA ($m=4, n=2$). The number of maximum attempts $t$ is set as 5. We randomly sample 300 examples for each dataset and conduct experiments on them. Statistically, the cost of our method is \$1.41 for 100 QA examples (about \textbf{\cent 1.4 per question}) and the inference time per example is about $40s$.\footnote{We acknowledge that our proposed method requires more cost compared with CoT or direct prompting. However, our approach is relatively cost-effective and the improved performance potentially leads to better health outcomes.}

\paragraph{Baselines.}
We have utilized models that are readily accessible through public APIs with the following baselines:

$\bullet$ \textbf{Setting w/o CoT}: 
Zero-shot \citep{kojima2022large} appends the prompt /emph{A: The answer is} to a given question and utilizes it as the input fed into LLMs. Few-shot \citep{few-shot} introduces several manually templated demonstrations, structured as $\left[ \text{Q: } \texttt{q}, \text{A: The answer is } \texttt{a}  \right]$, preceding the input question.

$\bullet$ \textbf{Setting w/ CoT}: 
Zero-shot CoT \citep{kojima2022large} directly incorporates the prompt \emph{Let's think step by step} after a question to facilitate inference. Few-shot CoT \citep{wei2022chain} adopts comparable methodologies to Few-shot but distinguishes itself by integrating rationales before deducing the answer.

$\bullet$ \textbf{Setting w/ SC}:
SC \citep{wang2022self} serves as an additional sampling method on Zero-shot CoT and Few-shot CoT, which yields the majority answer by sampling multiple chains.


\subsection{Main Results}\label{sec:ex_result}
Table \ref{tab:exp_result1} presents the main results on the nine datasets, including MedQA, MedMCQA, PubMedQA, and six subtasks from MMLU. We compare our method with several baselines in both zero-shot and few-shot settings. Notably, our proposed \textsc{MedAgents} framework outperforms the zero-shot baseline methods by a large margin, indicating the effectiveness of our \textsc{MedAgents} framework in real-world application scenarios. Furthermore, our approach achieves comparable performance under the zero-shot setting compared with the strong baseline \emph{Few-shot CoT+SC}. 

Interestingly, the introduction of CoT occasionally leads to a surprising degradation in performance.\footnote{A more detailed analysis of CoT's impact is provided in Appendix \ref{app:cot}.} We have found that reliance on CoT in isolation can inadvertently result in \emph{hallucinations} - spurious outputs typically associated with the misapplication of medical terminologies. In contrast, our multi-agent role-playing methodology effectively mitigates these issues, thus underscoring its potential as a more robust approach in medically oriented LLM applications.

\begin{table}[t]
	\centering
        \small
	\setlength{\tabcolsep}{15pt}
	{\begin{tabular}{ll}
		\toprule
		Method & Accuracy(\%) \\ 
		\midrule
            Direct Prompting  &  49.0 \\
            CoT Prompting  &  55.0 \\
            \cdashline{1-2}[1pt/2pt]
            \multicolumn{2}{l}{\textbf{w/ MedAgents}}\\
        
            \quad + Anal & 62.0($\uparrow 7.0$)   \\
            \quad + Anal \& Summ & 65.0($\uparrow 10.0$) \\
            \quad + Anal \& Summ \& Cons & 67.0($\uparrow 12.0$) \\
 
		\bottomrule
	\end{tabular}
	}
         \caption{Ablation study for different processes on MedQA. Anal: Analysis proposition, Summ: Report summarization, Cons: Collaborative consultation.}
	\label{tab:ablation}
\vspace{-.3cm}
\end{table}

\begin{table}[t]
	\centering
        \small
	\setlength{\tabcolsep}{5pt}
	{\begin{tabular}{lcc}
		\toprule
		\textbf{Method} & \textbf{MedQA} & \textbf{MedMCQA} \\ 
		\midrule
            \textsc{MedAgents} (GPT-3.5)  & 64.1 & 59.3\\
            \textsc{MedAgents} (GPT-4)    & 83.7 & 74.8 \\
            MedAlpaca-7B            & 55.2 & 45.8 \\
            BioMedGPT-10B           & 50.4 & 42.2 \\
            BioMedLM-2.7B           & 50.3 & - \\
            BioBERT (large)         & 36.7 & 37.1 \\
            SciBERT (large)         & - & 39.2 \\
            BERT (large)         & - & 33.6 \\
    	\bottomrule
	\end{tabular}
	}
         \caption{Comparison with open-source medical models.}
	\label{tab:comp}
         \vspace{-.4cm}
\end{table}

\begin{table}[t]
   \setlength{\tabcolsep}{3pt}
    \centering\small
    \setlength{\tabcolsep}{1pt}
\begin{tabular}{lcccc}
\toprule
 {Dataset} & {MedQA} & {MedMCQA} & {PubMedQA} & {MMLU}\\
 \midrule
  \#Question agents & {5} & {5} & {4} & {5}\\
  \#Option agents & {2} & {2} & {2} & {2}\\
\bottomrule
\end{tabular}
        \caption{Optimal number of agents on MedQA, MedMCQA, PubMedQA, and MMLU.\label{tab:num_agents}}
        \vspace{-.2cm}
\end{table}

\begin{table}[t]
	\centering
        \small
	\setlength{\tabcolsep}{5pt}
	{\begin{tabular}{lcc}
		\toprule
		\textbf{Method} & \textbf{MedQA} & \textbf{MedMCQA} \\ 
		\midrule
            \textsc{MedAgents}  &63.8  &58.9 \\
            Remove most relevant &60.5  &55.4 \\
            Remove least relevant  &66.2  &61.5 \\
            Remove randomly  &62.2  &56.3 \\
    	\bottomrule
	\end{tabular}
	}
         \caption{Domain variation study. The results are based on GPT-3.5.}
	\label{tab:domain_variation}
         \vspace{-.4cm}
\end{table}

\begin{figure}[t]
\centering
\includegraphics[width=0.45\textwidth]{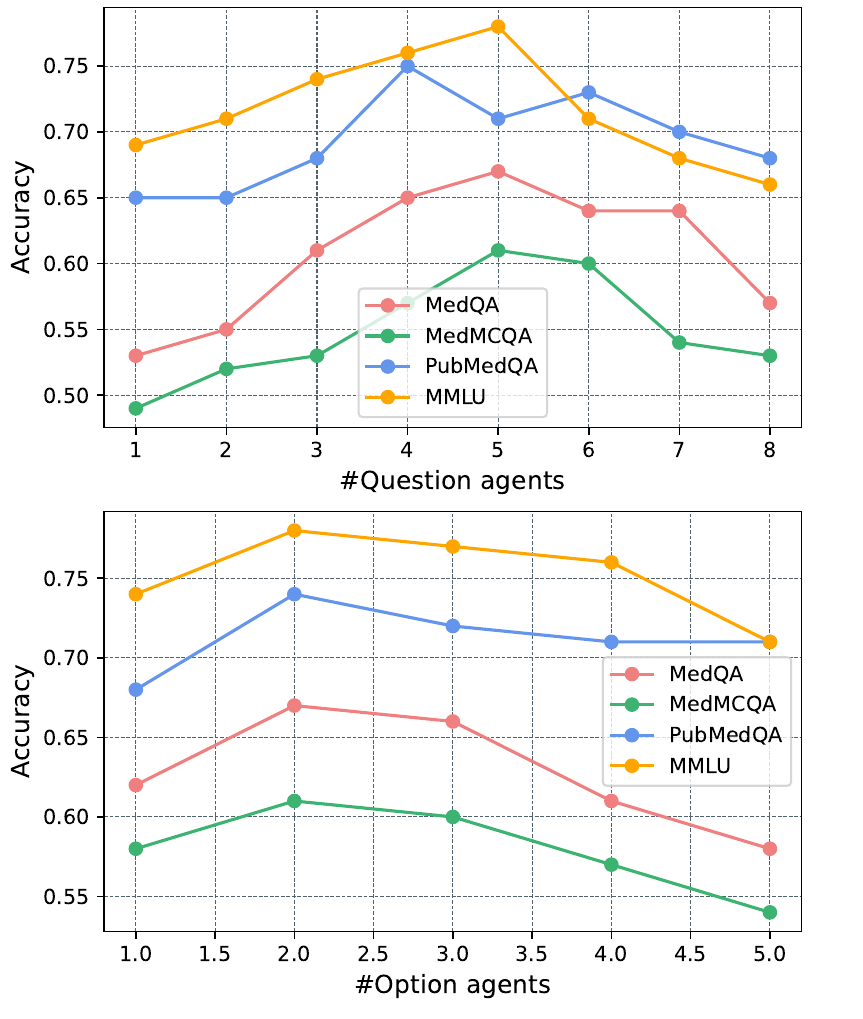}
\caption{Influence of the number of question and option agents on various datasets.}
\label{fig:num_agents}
\vspace{-.3cm}
\end{figure}


\section{Analysis}



\subsection{Ablation Study}
Since our \textsc{MedAgents} framework simulates a multi-disciplinary collaboration process that contains multiple intermediate steps, a natural question is whether each intermediate step contributes to the ultimate result. To investigate this, we ablate three major processes, namely \emph{analysis proposition}, \emph{report summarization} and \emph{collaborative consultation}. Results in Table \ref{tab:ablation} show that all of these processes are non-trivial. Notably, the proposition of \textsc{MedAgents} substantially boosts the performance (i.e., 55.0\%$\rightarrow$62.0\%), whereas the subsequent processes achieve relatively slight improvements over the previous one (i.e., 62.0\%$\rightarrow$65.0/67.0\%). 
This suggests that the initial role-playing agents are responsible for exploring medical knowledge of various levels and aspects within LLMs, while the following processes play a role in further verification and revision.

\subsection{Comparison with Open-source Medical Models}

We conduct a comprehensive comparison between our proposed \textsc{MedAgents} framework with more baseline methods, including open-source domain-adapted models such as MedAlpaca-7B \citep{han2023medalpaca}, BioMedGPT-10B \citep{luo2023biomedgpt}, BioMedLM-2.7B \citep{bolton2024biomedlm}, BioBERT (large) \citep{lee2020biobert}, SciBERT (large) \citep{beltagy2019scibert} and BERT (large).
We leverage them in the early stages of our work for preliminary attempts. Results in Table \ref{tab:comp} demonstrate that the open-source methods fell short of the baseline in Table \ref{tab:exp_result1}, which leads us to focus on the more effective methods.

\subsection{Number of agents}

As our \textsc{MedAgents} framework involves multiple agents that play certain roles to acquire the ultimate answer, we explore how the number of collaborating agents influences the overall performance. 
We vary the number of question and option agents while fixing other variables to observe the performance trends on the MedQA dataset. Figure \ref{fig:num_agents} and Table \ref{tab:num_agents} illustrate the corresponding trend and the optimal number of different agents. Our key observation lies in that the performance improves significantly with the introduction of any number of expert agents compared to our baseline, thus verifying the consistent contribution of multiple expert agents. We find that the optimal number of agents is relatively consistent across different datasets, pointing to its potential applicability to other datasets beyond those we test on.

\subsection{Domain Variation Study}

In order to investigate the influence of the changes in agent numbers, we perform additional studies where we manipulate agent numbers by selectively eliminating the most and least relevant domain experts based on domain relevance. Due to the manual evaluation involved in identifying the relevance of agent domains, our additional analysis was conducted on a limited set of 20 samples. Results in Table \ref{tab:domain_variation} depict minor variance for different sizes of data with random removing, reinforcing the notion that large-scale experiment performance remains largely robust against the effect of domain changes.

\subsection{Agent Quantity Study}
To further explore the effect of agent quantity without changes in domain representation, we conduct experiments with $k$ ($k=6$) identical-domain agents, then with $k-1$ and $k-2$, to observe performance shifts. The process of selecting these domains is automated via prompting, and our manual inspection confirms the high relevance and quality of the selected domains. The experiment is conducted on a dataset of 300 samples.

\begin{table}[t]
	\centering
        \small
	\setlength{\tabcolsep}{5pt}
	{\begin{tabular}{lcc}
		\toprule
		\textbf{Method} & \textbf{MedQA} & \textbf{MedMCQA} \\ 
		\midrule
            \multicolumn{3}{l}{\textbf{\textsc{MedAgents}}}\\
        
            \quad w/ 6 different domains &64.1  &59.3   \\
            \quad w/ 6 same domains &59.2  &58.1   \\
            \quad w/ 5 same domains &57.5  &57.3   \\
            \quad w/ 4 same domains &55.9  &57.0   \\

		\bottomrule
	\end{tabular}
	}
         \vspace{-.2cm}
         \caption{Agent quantity study. The results are based on GPT-3.5.}
	\label{tab:agent_quantity}
 \vspace{-.4cm}
\end{table}

\subsection{Error Analysis}

Based on our results, we conduct a human evaluation to pinpoint the limitations and issues prevalent in our model. We distill these errors into four major categories:
(i) \textbf{Lack of Domain Knowledge}: the model demonstrates an inadequate understanding of the specific medical knowledge necessary to provide an accurate response;
(ii) \textbf{Mis-retrieval of Domain Knowledge}: the model has the necessary domain knowledge but fails to retrieve or apply it correctly in the given context;
(iii) \textbf{Consistency Errors}: the model provides differing responses to the same statement. The inconsistency suggests confusion in the model's understanding or application of the underlying knowledge;
(iv) \textbf{CoT Errors}: the model may form and follow inaccurate rationales, leading to incorrect conclusions.

\begin{figure*}[t]
\centering
\includegraphics[width=1.0\textwidth]{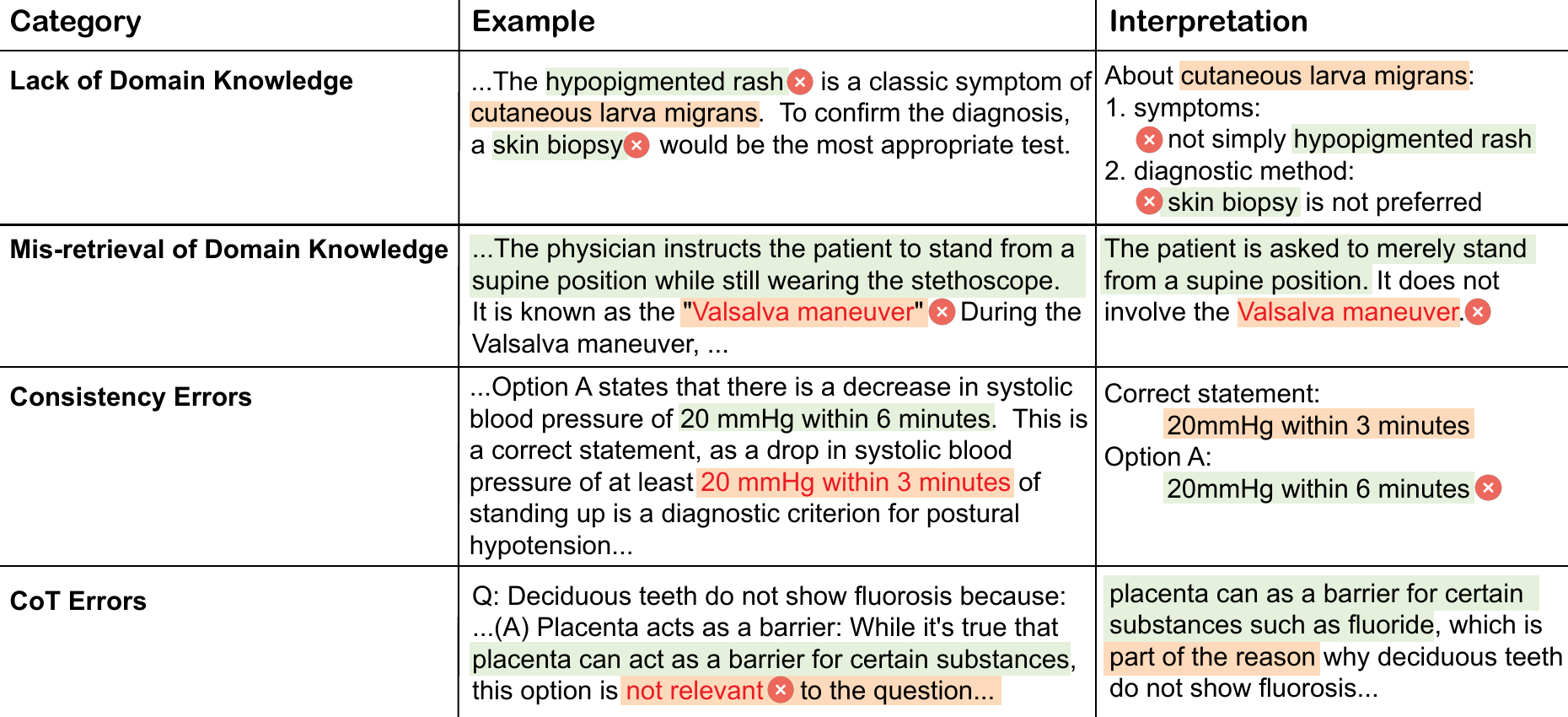}
        \vspace{-.1cm}
\vspace{-.1cm}
\vspace{-.2cm}

\caption{Examples of error cases from MedQA and MedMCQA datasets in four major categories including lack of domain knowledge, mis-retrieval of domain knowledge, consistency errors, and CoT errors.}
\label{fig:wrong_cases}
\vspace{-.4cm}
\end{figure*}

To illustrate the error examples intuitively, we select four typical samples from the four error categories, which can be shown in Figure \ref{fig:wrong_cases}:
(i) The first error is due to a lack of domain knowledge regarding \emph{cutaneous larva migrans}, whose symptoms are not purely \emph{hypopigmented rash}, as well as the fact that \emph{skin biopsy} is not an appropriate test method, which results in the hallucination phenomenon.
(ii) The second error is caused by mis-retrieval of domain knowledge, wherein the fact in green is not relevant to \emph{Valsalva maneuver}.
(iii) The third error is attributed to consistency errors, where the model incorrectly regards \emph{20 mmHg within 6 minutes} and \emph{20 mmHg within 3 minutes} as the same meaning.
(iv) The fourth error is provoked by incorrect inference about the relevance of a fact and option A in CoT.

\begin{figure}[t]
\centering
\includegraphics[width=0.34\textwidth]{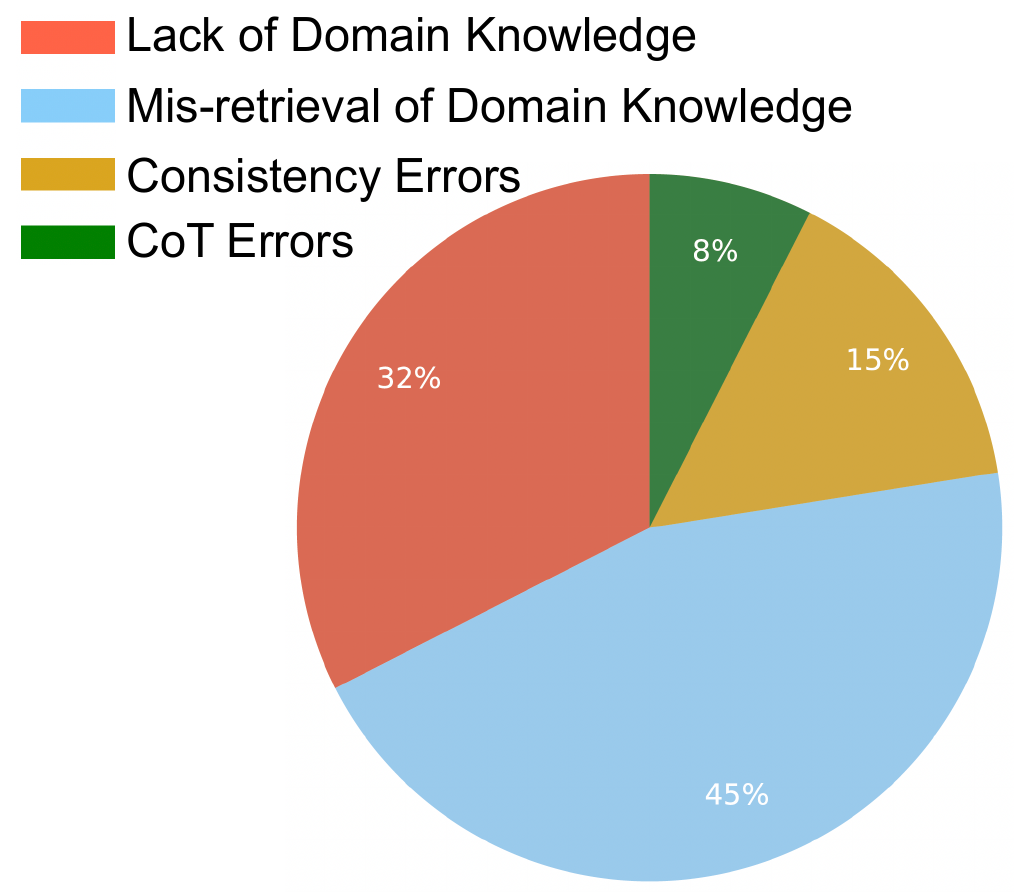}
\caption{Ratio of different categories in error cases.}
\label{fig:error}
\vspace{-.5cm}
\end{figure}

Furthermore, we analyze the percentage of different categories by randomly selecting 40 error cases in MedQA and MedMCQA datasets. As is shown in Figure \ref{fig:error}, the majority (77\%) of the error examples are due to confusion about the domain knowledge (including the lack and mis-retrieval of domain knowledge), which illustrates that there still exists a portion of domain knowledge that is explicitly beyond the intrinsic knowledge of LLMs, leading to a bottleneck of our proposed method. As a result, our analysis sheds light on future directions to mitigate the aforementioned drawbacks and further strengthen the model's proficiency and reliability. 



\subsection{Correctional Capabilities and Interpretability}
In our extensive examination of the \textsc{MedAgents} framework, we discover the decent correctional capabilities of our framework. Please refer to Appendix \ref{app:case} and Table \ref{tab:case1} for an in-depth overview of instances where our approach successfully amends previous inaccuracies, steering the discussion towards more accurate outcomes. These corrections showcase the \textsc{MedAgents} framework's strength in collaborative synthesis; it distills and integrates diverse expert opinions into a cohesive and accurate conclusion. By interweaving a variety of perspectives, the collaborative consultation actively refines and rectifies initial analyses, thereby aligning the decision-making process closer to clinical accuracy. This iterative refinement serves as a practical demonstration of our model's proficiency in rectifying errors, substantiating its interpretability and accuracy in complex medical reasoning tasks.

\section{Related Work}

\subsection{LLMs in Medical Domains}
Recent years have seen remarkable progress in the application of LLMs \citep{wu2023precedentenhanced, clinical, yang2023investlm}, with a particularly notable impact on the medical field \citep{bao2023discmedllm, nori2023capabilities, rosol2023evaluation}. Although LLMs have demonstrated their potential in distinct medical applications encompassing diagnostics \citep{clinical,han2023medalpaca}, genetics \citep{duong2023analysis, jin2023genegpt}, pharmacist \citep{liu2023pharmacygpt}, and medical evidence summarization \citep{tang2023aligning, tang2023evaluating,shaib2023summarizing}, concerns persist when LLMs encounter clinical inquiries that demand intricate medical expertise and decent reasoning abilities \citep{umapathi2023med, clinical}. Thus, it is of crucial importance to further arm LLMs with enhanced clinical reasoning capabilities. Currently, there are two major lines of research on LLMs in medical domains, tool-augmented methods and instruction-tuning methods.

For tool-augmented approaches, recent studies rely on external tools to acquire additional information for clinical reasoning. For instance, GeneGPT \citep{jin2023genegpt} guided LLMs to leverage the Web APIs of the National Center for Biotechnology Information (NCBI) to meet various biomedical information needs. \citet{zakka2023almanac} proposed Almanac, a framework that is augmented with retrieval capabilities for medical guidelines and treatment recommendations. 
\citet{kang2023knowledge} introduced a method named KARD to improve small LMs on specific domain knowledge by fine-tuning small LMs on the rationales generated from LLMs and augmenting small LMs with external knowledge from a non-parametric memory.

Current instruction tuning research predominantly leverages external clinical knowledge bases and self-prompted data to obtain instruction datasets \citep{tu2023towards, zhang2023pmc, singhal2023towards, tang2023gersteinlab}. 
These datasets are then employed to fine-tune LLMs within the medical field \citep{singhal2023towards}. 
Some of these models utilize a wide array of datasets collected from medical and biomedical literature, fine-tuned with specialized or open-ended instruction data \citep{li2023llava,singhal2023towards}. Others focus on specific areas such as traditional Chinese medicine or large-scale, diverse medical instruction data to enhance their medical proficiency \citep{tan2023medchatzh, zhang2023alpacareinstructiontuned}. 
Unlike these methods, our work emphasizes harnessing latent medical knowledge intrinsic to LLMs and improving reasoning in a training-free setting.

\subsection{LLM-based Multi-agent Collaboration}
The development of LLM-based agents has made significant progress in the community by endowing LLMs with the ability to perceive surroundings and make decisions individually \citep{wang2023survey,yao2022react,nakajima2023task,xie2023openagents,zhou2023webarena}. Beyond the initial single-agent mode, the multi-agent pattern has garnered increasing attention recently \citep{xi2023rise, li2023metaagents, hong2023metagpt} which further explores the potential of LLM-based agents by learning from multi-turn feedback and cooperation.
In essence, the key to LLM-based multi-agent collaboration is the simulation of human activities such as role-playing \citep{wang2023unleashing, hong2023metagpt} and communication \citep{wu2023autogen, qian2023communicative, li2023camel, li2023theory}.

For instance, Solo Performance Prompting (SPP) \citep{wang2023unleashing} managed to combine the strengths of multiple minds to improve performance by dynamically identifying and engaging multiple personas throughout task-solving. Camel \citep{li2023camel} leveraged role-playing to enable chat agents to communicate with each other for task completion. 

Several recent works attempt to incorporate adversarial collaboration including debates \citep{du2023improving, xiong2023examining} and negotiation \citep{fu2023improving} among multiple agents to further boost performance. 
\citet{liang2023encouraging} proposed a multi-agent debate framework in which various agents put forward their statements in a \emph{tit for tat} pattern. Inspired by the multi-disciplinary consultation mechanism which is common and effective in hospitals, we are thus inspired to apply this mechanism to medical reasoning tasks through LLM-based multi-agent collaboration.

\section{Conclusion}
We present a novel medical QA framework that uses role-playing agents for multi-round discussions, offering greater reliability and clarity without prior training. Our method surpasses zero-shot baselines and matches few-shot baselines across nine datasets. Despite successes, human-based evaluations of errors have highlighted areas for refinement. Our approach differs from most existing methods by eliminating the dependency on knowledge bases, instead uniquely integrating medical knowledge through role-playing agents.

\newpage
\section*{Limitation}
The proposed \textsc{MedAgents} framework has shown promising results, but there are still a few points that could be addressed in future studies.
First, the parameterized knowledge within LLMs may need updating over time, and thus, continuous efforts are required to keep the framework up-to-date.
Second, integrating diverse models at different stages of our framework might be an intriguing exploration.
Third, the framework may have limited applicability in low-resource languages. Adapting this framework to a wider range of low-resource languages could meet their specific medical needs to some extent.

\section*{Ethics Statement}
Although our work strictly adheres to well-established benchmarks in the field of medical question answering, it is possible that our approach introduces potential risks, e.g., some inherent biases of LLMs, when applying LLM reasoning to critical areas such as medicine.

\section*{Acknowledgments}
Xiangru Tang and Mark Gerstein are supported by Schmidt Sciences. Zhuosheng Zhang is supported by CIPSC-SMP-Zhipu.AI Large Model Cross-Disciplinary Fund.

\bibliography{anthology,custom}

\newpage

\appendix

\clearpage

\section{Analysis on the Addition of CoT}\label{app:cot}

\begingroup
\begin{table*}[!]
    \centering
    \caption{
    Examples showing the influence of the addition of CoT \label{tab:cot-examples}
    }
    \vspace{2.8mm}
    \begin{tabular}{p{0.96\linewidth}}
        \toprule
        \textbf{\texttt{Question:}} Which of the following is the large bone found superior to the patella and inferior to the ischium? (A) Calcaneus (B) Femur (C) Symphysis pubis (D) Tibia \\
        \textbf{\texttt{Zero-shot Answer:}} B \okmark\\
        \textbf{\texttt{Zero-shot-CoT Answer:}} D \ngmark\\
        \textbf{\texttt{Zero-shot-CoT Rationale:}} \textcolor{blue}{The bone found superior to the patella and inferior to the ischium is most likely the tibia} (\ngmark). Let's go through the options to confirm. Option A: Calcaneus - The calcaneus is the heel bone and is not located in the specified position. Option B: Femur - \textcolor{blue}{The femur is the thigh bone and is located superior to the patella, but it is not inferior to the ischium} (\ngmark). Option C: Symphysis pubis - The symphysis pubis is a joint located in the pelvis and is not a bone. Option D: Tibia - The tibia is the shin bone and is located superior to the patella and inferior to the ischium. This seems to be the correct option. Answer: D\\
        

\bottomrule
\end{tabular}
\end{table*}
\endgroup

We provide an intriguing example that reveals a seemingly counter-intuitive observation: the addition of the CoT in a zero-shot setting led to a performance drop compared to the zero-shot one.

As demonstrated in the example, for specialist domains that demand considerable expert knowledge such as the medical domain knowledge, employing a CoT approach might sometimes lead to \emph{hallucination} \citep{bubeck2023sparks, guerreiro2023hallucinations, ji2023survey, maynez2020faithfulness}. \emph{Hallucinations} refer to instances where the language model starts generating inaccurate or irrelevant information based on its insufficient understanding \citep{wei2022chain, kojima2022large, shi2023large}. Consequently, in these instances, the use of the CoT method does not improve but hindrances the overall performance.

This issue is particularly stated in the medical question-answering field by some recent work, where it has been demonstrated that the CoT's step-by-step approach is unable to generate correct medical answers effectively. For example, 
the results from \citet{lievin2022can} demonstrate that CoT improvements are significantly limited.

Such failures in medical question-answering originate from a lack of domain knowledge \citep{harris2023large, kung2023performance, tian2024opportunities} instead of reasoning rationale. This was also observed in our experiments, with a substantial 77\% of errors related to domain knowledge, compared to a minor 8\% due to CoT (as shown in Figure \ref{fig:error}).

To address these issues, there has indeed been a recent shift toward utilizing RAG (Retrieval-Augmented Generation) for domain knowledge enhancement \citep{wang2023augmenting, lala2023paperqa} in medical QA. Despite these developments, our study underscores a novel approach within this field: our use of role-playing. Role-playing in our \textsc{MedAgents} framework allows the model to explicitly reason with accurate knowledge, remarkably bypassing the need for RAG.

\clearpage

\section{Interpretability: Case Study}\label{app:case}
We analyze instances where our proposed \textsc{MedAgents} framework manages to correct previous errors. Table \ref{tab:case1} illustrates examples of our approach to turning previously incorrect answers into correct ones. 
These cases showcase the robustness of our \textsc{MedAgents} framework. Different medical professionals offer depth and breadth in the analysis, transforming it into a comprehensive assessment. Furthermore, the collaborative consultation refines and corrects the initial interpretation, leading us toward the correct answer, thus demonstrating the effectiveness of our method in correcting errors and ensuring accuracy.

\begingroup\label{tab:case1}
\begin{table*}[!]
    \centering
    \caption{Case studies demonstrating how our proposed \textsc{MedAgents} framework manages to correct previous errors.
    }
    \vspace{2.8mm}
    \begin{tabular}{p{0.96\linewidth}}
        \toprule
        \sethlcolor{mygreen}\hl{\textbf{\texttt{Question:}}} A 23-year-old college student experiences monthly chest discomfort, shortness of breath, shakiness, and excessive sweating, likely due to stress. He fears having an episode in public, causing him to avoid leaving home. His medical history is unclear, and the physical exam is normal. Which of the following is the best medication for the long-term management of this patient's condition? (A) Citalopram (B) Lithium (C) Lorazepam (D) Propranolol (E) Quetiapine \\

        \\
        
        \sethlcolor{mygreen}\hl{\textbf{\texttt{Experts:}}} Cardiology, Psychiatry, Pulmonology, Endocrinology, Neurology\\

        \\
        
        \sethlcolor{mygreen}\hl{\textbf{\texttt{Analyses from various experts:}}}
        \\
        \quad \textbf{\underline{Cardiology Expert}}: ...These symptoms are not typical of a cardiac cause given the patient's young age and normal physical exam...The best medication for long-term management of anxiety disorders is often selective serotonin reuptake inhibitors (SSRIs) like \textcolor{blue}{Citalopram}.
        \\
        \quad \textbf{\underline{Psychiatry Expert}}: ...Although \textcolor{blue}{quetiapine} is commonly used in the treatment of schizophrenia and bipolar disorder, it is also effective in controlling the symptoms of rare endocrine disorders that present with anxiety-like symptoms.
        \\
        \quad \textbf{\underline{Pulmonology Expert}}: ...this treatment based on the possibility that the symptoms could be due to a respiratory condition, such as hyperventilation syndrome, which \textcolor{blue}{Propranolol} is thought to help by reducing the heart rate and perceived shortness of breath.
        \\
        \quad \textbf{\underline{Endocrinology Expert}}: SSRIs like \textcolor{blue}{Citalopram} can affect endocrine function by regulating serotonin levels, which in turn can modulate the hypothalamic-pituitary-adrenal (HPA) axis...
        \\
        \quad \textbf{\underline{Neurology Expert}}: the focus would be on addressing the dysregulated ANS activity that is causing the physical symptoms of stress. The most appropriate medication would likely be \textcolor{blue}{Citalopram}, an SSRI, which can help improve mood and reduce anxiety...
        \\
        \quad \textcolor{magenta}{Expert analysis presents a multifaceted view of the problem. Based on the experts' specialties, their analyses focus on different aspects, leading to the suggestion of two drugs, \textbf{Citalopram and Propranolol}.}
        
        \\
        
        \sethlcolor{mygreen}\hl{\textbf{\texttt{Synthesized Report (v0):}}} ... Based on the suggestions from the experts, the medication should be \textcolor{blue}{Citalopram} along with \textcolor{blue}{Propranolol}, which can help alleviate shortness of breath...
        \\
        \quad \textcolor{magenta}{The first synthesis suggested a combination of Citalopram and Propranolol.}
        
        \\

        \sethlcolor{mygreen}\hl{\textbf{\texttt{Vote Result:}}}  
         \\
        \quad \textbf{\underline{Cardiology Expert}}: No
        \\
        \quad \textbf{\underline{Psychiatry Expert}}: Yes
        \\
        \quad \textbf{\underline{Pulmonology Expert}}: Yes
        \\
        \quad \textbf{\underline{Endocrinology Expert}}: Yes
        \\
        \quad \textbf{\underline{Neurology Expert}}: No
        
        \\
        
        \sethlcolor{mygreen}\hl{\textbf{\texttt{Synthesized Report (v1):}}} ...the most appropriate medication for the long-term management of this patient's condition appears to be \textcolor{blue}{Citalopram}(SSRI).\\
        \quad
        \textcolor{magenta}{The final report suggested \textbf{Citalopram} as the best medication for the patient's condition.} 
        
        \\


\bottomrule
\end{tabular}
\end{table*}

\begin{table*}[!]
    \centering
    \caption{Case studies demonstrating how our proposed \textsc{MedAgents} framework manages to correct previous errors.
    }
    \vspace{2.8mm}
    \begin{tabular}{p{0.96\linewidth}}
        \toprule
        \sethlcolor{mygreen}\hl{\textbf{\texttt{Question:}}} Eustachian tube function is best assessed by: (A) Tympanometry (B) VEMP (C) Rhinomanometry (D) Politzer tests \\

        \\
        
        \sethlcolor{mygreen}\hl{\textbf{\texttt{Experts:}}} Otolaryngology, Audiology, Radiology, Allergy and Immunology, Pulmonology\\

        \\
        
        \sethlcolor{mygreen}\hl{\textbf{\texttt{Analyses from various experts:}}}
        \\
        \quad \textbf{\underline{Otolaryngology Expert}}: 
        ...The focus is on Eustachian tube function, which is essential for equalizing ear pressure and facilitating normal hearing. Several tests are available, but \textcolor{blue}{tympanometry} is a primary method for assessing function, indicating if there's a blockage or proper movement of the eardrum.
        \\
        \quad \textbf{\underline{Audiology Expert}}: 
        ...\textcolor{blue}{Tympanometry} is correct, VEMP is not related, Rhinomanometry is unrelated, and \textcolor{blue}{the Politzer test} is appropriate for Eustachian tube assessment.
        \\
        \quad \textbf{\underline{Radiology Expert}}: ...Proper function of the Eustachian tube is vital for hearing and balance. \textcolor{blue}{Tympanometry} is one of the methods used to assess this function, by evaluating the compliance of the middle ear.
        \\
        \quad \textbf{\underline{Allergy and Immunology Expert}}: 
        Assessing Eustachian tube function is crucial for diagnosing related dysfunctions and infections. \textcolor{blue}{Tympanometry} is a valuable test to determine if the tube is functioning correctly.
        \\
        \quad \textbf{\underline{Pulmonology Expert}}: ...The assessment of the Eustachian tube function is important for diagnosing ear conditions that could affect breathing. \textcolor{blue}{Tympanometry} helps evaluate eardrum mobility and middle ear pressure
        \\
        \quad \textcolor{magenta}{Based on the experts' specialties, their analyses focus on different aspects, leading to the suggestion of two methods, \textbf{Tympanometry and Politzer test}.}
        
        \\
        
        \sethlcolor{mygreen}\hl{\textbf{\texttt{Synthesized Report (v0):}}} ...\textcolor{blue}{Tympanometry} is reaffirmed as a method to assess Eustachian tube function, with additional methods like sonotubometry and nasal endoscopy also important. The \textcolor{blue}{Politzer test} is recognized as another method for such assessments.
        \\
        \quad \textcolor{magenta}{The first synthesis suggested \textbf{Tympanometry and Politzer test}.}
        
        \\

        \sethlcolor{mygreen}\hl{\textbf{\texttt{Vote Result:}}}  
         \\
        \quad \textbf{\underline{Otolaryngology Expert}}: Yes
        \\
        \quad \textbf{\underline{Audiology Expert}}: Yes
        \\
        \quad \textbf{\underline{Radiology Expert}}: Yes
        \\
        \quad \textbf{\underline{Allergy and Immunology Expert}}: Yes
        \\
        \quad \textbf{\underline{Pulmonology Expert}}:  Initially No, then Yes after revision.
        
        \\
        
        \sethlcolor{mygreen}\hl{\textbf{\texttt{Synthesized Report (v1):}}} ...\textcolor{blue}{Tympanometry} is a key method among several to assess Eustachian tube function, crucial for diagnosing Eustachian tube dysfunction, infections, and hearing issues....\\
        \quad
        \textcolor{magenta}{The final report suggested \textbf{Tympanometry} as the best medication for the patient's condition.} \\
\bottomrule
\end{tabular}
\end{table*}

\begin{table*}[!]
    \centering
    \caption{Case studies demonstrating how our proposed \textsc{MedAgents} framework manages to correct previous errors.
    }
    \vspace{2.8mm}
    \begin{tabular}{p{0.96\linewidth}}
        \toprule
        \sethlcolor{mygreen}\hl{\textbf{\texttt{Question:}}} 
        To prevent desiccation and injury, the embryos of terrestrial vertebrates are encased within a fluid secreted by the? (A) amnion (B) chorion (C) allantois (D) yolk sac
        \\

        \\
        
        \sethlcolor{mygreen}\hl{\textbf{\texttt{Experts:}}} Embryology, Physiology, Dermatology, Endocrinology, Reproductive Medicine\\

        \\
        
        \sethlcolor{mygreen}\hl{\textbf{\texttt{Analyses from various experts:}}}
        \\
        \quad \textbf{\underline{Embryology Expert}}: 
        ... \textcolor{blue}{amnion} is the primary structure responsible for secreting amniotic fluid, with the \textcolor{blue}{chorion} also contributing to this process.
        \\
        \quad \textbf{\underline{Physiology Expert}}: 
        ...the amniotic sac and amniotic fluid, produced by the \textcolor{blue}{amnion}, is important in creating a stable and protected environment for the embryo. Besides, its cushioning effect and role in the exchange of nutrients and gases cannot be ignored either.
        \\
        \quad \textbf{\underline{Dermatology Expert}}: 
        ...the importance of \textcolor{blue}{amniotic} fluid in fetal development, particularly for the skin...
        \\
        \quad \textbf{\underline{Endocrinology Expert}}: 
        the question relates to embryological development rather than an endocrinological condition, ... the \textcolor{blue}{amniotic} sac plays a vital role in providing a protective environment.
        \\
        \quad \textbf{\underline{Reproductive Medicine Expert}}: 
        ...\textcolor{blue}{amniotic} fluid is of crucial importance for the normal development and survival of the embryo.
        \\
        \quad \textcolor{magenta}{Based on the experts' specialties, their analyses focus on different aspects, leading to the suggestion of two methods, \textbf{amnion and chorion}.}
        
        \\
        
        \sethlcolor{mygreen}\hl{\textbf{\texttt{Synthesized Report (v0):}}} ... Both the \textcolor{blue}{amnion} and \textcolor{blue}{chorion} play vital roles in this process.
        \\
        \quad \textcolor{magenta}{The first synthesis suggested \textbf{amnion and chorion}.}
        
        \\

        \sethlcolor{mygreen}\hl{\textbf{\texttt{Vote Result:}}}  
         \\
        \quad \textbf{\underline{Embryology Expert}}: Yes
        \\
        \quad \textbf{\underline{Physiology Expert}}: Yes
        \\
        \quad \textbf{\underline{Dermatology Expert}}: No
        \\
        \quad \textbf{\underline{Endocrinology Expert}}: No
        \\
        \quad \textbf{\underline{Reproductive Medicine Expert}}: Yes
        
        \\
        
        \sethlcolor{mygreen}\hl{\textbf{\texttt{Synthesized Report (v1):}}} ... the \textcolor{blue}{amnion} as the \underline{primary source} of the amniotic fluid that protects the developing embryos of terrestrial vertebrates, with an understanding of the contributions from other structures like the chorion and the allantois.\\
        \quad
        \textcolor{magenta}{Dermatology and Endocrinology experts have revised their initial focus to align with the consensus on the role of the amnion in secreting amniotic fluid.} \\
\bottomrule
\end{tabular}
\end{table*}

\endgroup

\clearpage

\section{Dataset Information}\label{app:dataset}
MedQA consists of USMLE-style questions with four or five possible answers. MedMCQA encompasses four-option multiple-choice questions from Indian medical entrance examinations (AIIMS/NEET). MMLU (Massive Multitask Language Understanding) covers 57 subjects across various disciplines, including STEM, humanities, social sciences, and many others. 
The scope of its assessment stretches from elementary to advanced professional levels, evaluating both world knowledge and problem-solving capabilities. While the subject areas tested are diverse, encompassing traditional fields like mathematics and history, as well as more specialized areas like law and ethics, we deliberately limit our selection to the sub-subjects within the medical domain for this exercise, following ~\citep{clinical}.

\clearpage

\section{Prompt Templates}\label{app:prompt}
Prompt templates involved in the experiments are presented in Table \ref{tab:prompt}.

\begingroup
\begin{table*}[!]
    \centering
    \caption{
    Prompt templates and role descriptions employed in our \textsc{MedAgents} framework. \label{tab:prompt}
    }
    \vspace{-3mm}
    \begin{tabular}{p{0.96\linewidth}}
        \toprule
        \textbf{$\texttt{r}_\texttt{qd}$:} You are a medical expert who specializes in categorizing a specific medical scenario into specific areas of medicine. \\
        \textbf{$\texttt{prompt}_\texttt{qd}$:} You need to complete the following steps: 1. Carefully read the medical scenario presented in the question: \texttt{question}. 2. Based on the medical scenario in it, classify the question into five different subfields of medicine. 3. You should output in the same format as: \texttt{Medical Field: | }. \\
        
        \vspace{-2.5mm}
        \textbf{$\texttt{r}_\texttt{od}$:} As a medical expert, you possess the ability to discern the two most relevant fields of expertise needed to address a multiple-choice question encapsulating a specific medical context. \\
        \textbf{$\texttt{prompt}_\texttt{od}$:} You need to complete the following steps: 1. 1. Carefully read the medical scenario presented in the question: \texttt{question}. 2. The available options are: \texttt{options}. Strive to understand the fundamental connections between the question and the options. 3. Your core aim should be to categorize the options into two distinct subfields of medicine. You should output in the same format as: \texttt{Medical Field: | }. \\

        \vspace{-2.5mm}
        \textbf{$\texttt{r}_\texttt{qa}$:} You are a medical expert in the domain of \texttt{question\_domain}. From your area of specialization, you will scrutinize and diagnose the symptoms presented by patients in specific medical scenarios.\\
        \textbf{$\texttt{prompt}_\texttt{qa}$:} Please meticulously examine the medical scenario outlined in this question: \texttt{question}. Drawing upon your medical expertise, interpret the condition being depicted. Subsequently, identify and highlight the aspects of the issue that you find most alarming or noteworthy. \\

        \vspace{-2.5mm}
        \textbf{$\texttt{r}_\texttt{oa}$:} You are a medical expert specialized in the \texttt{op\_domain} domain. You are adept at comprehending the nexus between questions and choices in multiple-choice exams and determining their validity. Your task, in particular, is to analyze individual options with your expert medical knowledge and evaluate their relevancy and correctness.\\
        \textbf{$\texttt{prompt}_\texttt{oa}$:} Regarding the question: \texttt{question}, we procured the analysis of five experts from diverse domains. The evaluation from the \texttt{question\_domain} expert suggests: \texttt{question\_analysis}. The following are the options available: \texttt{options}. Reviewing the question's analysis from the expert team, you're required to fathom the connection between the options and the question from the perspective of your respective domain and scrutinize each option individually to assess whether it is plausible or should be eliminated based on reason and logic. Pay close attention to discerning the disparities among the different options and rationalize their existence. A handful of these options might seem right at first glance but could potentially be misleading in reality. \\

        \vspace{-2.5mm}
        \textbf{$\texttt{r}_\texttt{rs}$:} You are a medical assistant who excels at summarizing and synthesizing based on multiple experts from various domain experts.\\
        \textbf{$\texttt{prompt}_\texttt{rs}$:} Here are some reports from different medical domain experts. You need to complete the following steps: 1. Take careful and comprehensive consideration of the following reports. 2. Extract key knowledge from the following reports. 3. Derive the comprehensive and summarized analysis based on the knowledge. 4. Your ultimate goal is to derive a refined and synthesized report based on the following reports. You should output in exactly the same format as: \texttt{Key Knowledge:; Total Analysis:} \\

        \vspace{-2.5mm}
        \textbf{$\texttt{r}_\texttt{vote}$:} You are a medical expert specialized in the \texttt{domain} domain.\\
        \textbf{$\texttt{prompt}_\texttt{vote}$:} Here is a medical report: \texttt{synthesized\_report}. As a medical expert specialized in \texttt{domain}, please carefully read the report and decide whether your opinions are consistent with this report. Please respond only with: \texttt{[YES or NO]}. \\

        \vspace{-2.5mm}
        \textbf{$\texttt{prompt}_\texttt{mod}$:} Here is advice from a medical expert specialized in \texttt{domain}: \texttt{advice}. Based on the above advice, output the revised analysis in the same format as: \texttt{Key Knowledge:; Total Analysis:} \\

        \vspace{-2.5mm}
        \textbf{$\texttt{prompt}_\texttt{dm}$:} Here is a synthesized report: \texttt{syn\_report}. Based on the above report, select the optimal choice to answer the question. Points to note: 1. The analyses provided should guide you towards the correct response. 2. Any option containing incorrect information inherently cannot be the correct choice. 3. Please respond only with the selected option's letter, like A, B, C, D, or E, using the following format: \texttt{'''Option: [Selected Option's Letter]'''}. Remember, it's the letter we need, not the full content of the option. \\
\bottomrule
\end{tabular}
\end{table*}
\endgroup

\clearpage
\section{Multiple Runs}
we have performed multiple runs on a set of 300 samples for GPT-4 and GPT-3.5 to account for variability.

The preliminary tests involved the average score of 5 repetitions for each sample, and the results indicated a decent small variance between runs. In summary, the consistency observed in this set provides confidence in the stability of our reported results.

\begin{table}[h]
\centering
\begin{tabular}{l l l}
\toprule
 & MedQA & MedMCQA \\ 

MC framework GPT-3.5 (single try) & 64.1 & 59.3 \\ 
MC framework GPT-3.5 (5 repetitions) & 64.3 & 59.2 \\
MC framework GPT-4 (single try) & 83.7 & 74.8 \\
MC framework GPT-4 (5 repetitions) & 83.5 & 74.9 \\
\bottomrule
\end{tabular}
\caption{GPT-4 versus GPT-3.5 based on multiple runs of 300 samples.}
\label{table:your_label}
\end{table}

\end{document}